\documentclass{article}
\usepackage{ijcai25}

\usepackage{amssymb}
\usepackage{times}
\usepackage{soul}
\usepackage{url}
\usepackage[hidelinks]{hyperref}
\usepackage[utf8]{inputenc}
\usepackage[small]{caption}
\usepackage{graphicx}
\usepackage{amsmath}
\usepackage{amsthm}
\usepackage{booktabs}
\usepackage{algorithm}
\usepackage{algorithmic}
\usepackage[switch]{lineno}
\usepackage{multirow}
\usepackage{multicol}
\usepackage{xcolor}
\usepackage{booktabs}
\usepackage{multirow}
\usepackage{adjustbox}

\newtheorem{definition}{Definition}

\title{GRAML: Dynamic Goal Recognition As Metric Learning}

\author{
Matan Shamir$^1$
\and
Reuth Mirsky$^{1,2}$ \\
\affiliations
$^1$Computer Science Department, Bar-Ilan University, Israel\\
$^2$Computer Science Department, Tufts University, MA, USA\\
\emails
matan.shamir@live.biu.ac.il,
reuth.mirsky@tufts.edu
}

\begin{document}

\maketitle

\begin{abstract}
Goal Recognition (GR) is the problem of recognizing an agent's objectives based on observed actions. 
Recent data-driven approaches for GR alleviate the need for costly, manually crafted domain models. 
However, these approaches can only reason about a pre-defined set of goals, and time-consuming training is needed for new emerging goals. 
To keep this model-learning automated while enabling quick adaptation to new goals, this paper introduces GRAML: Goal Recognition As Metric Learning. 
GRAML uses a Siamese network to treat GR as a deep metric learning task, employing an RNN that learns a metric over an embedding space, where the embeddings for observation traces leading to different goals are distant, and embeddings of traces leading to the same goals are close.
This metric is especially useful when adapting to new goals, even if given just one example observation trace per goal. 
Evaluated on a versatile set of environments, GRAML shows speed, flexibility, and runtime improvements over the state-of-the-art GR while maintaining accurate recognition.
\end{abstract}

\section{Introduction}
Goal Recognition (GR) involves deducing an agent's objective based on observed actions. It plays a crucial role in various settings, including Human-Robot Interaction~\cite{massardi2020parc,shvo2022proactive} and Multi-Agent Systems~\cite{avrahami2005fast,freedman2017integration,su2023fast}. GR problem formulations typically assume a fixed set of goals~\cite{asai2022classical,mirsky2021symbolic,RamirezGeffner09,Pereira_Meneguzzi_2023,ramirez2011goal}. 
However, consider a service robot recognizing which dish a person is making out of a recipe book with a finite set of recipes. 
If the robot finds that a recipe was added to the book, it will need to form a new GR problem containing the new recipe and adjust its internal state, a potentially costly process, as traditional GR methods assume a pre-defined set of goals. Adapting to the new goal and performing GR is part of a single problem named Online Dynamic Goal Recognition (ODGR). In ODGR, similar, consequent GR problems share identical or similar domain descriptions but vary in their set of goals \cite{shamir2024RLC}.
The paper introduces a new ODGR framework called \textit{Goal Recognition as Metric Learning (GRAML)}, which aims for a one-shot solution when a new set of goals emerges, supporting both discrete and continuous domains. In line with other learning-based GR approaches, GRAML involves a distinct \textit{learning phase} where the domain theory is constructed and an \textit{inference phase} where observations are given. In addition to these phases, GRAML addresses the potential shift in the set of potential goals between GR problems. It does so by incorporating an \textit{adaptation phase}, during which the framework adjusts its internal state to align with the newly emerged goals. 
%
GRAML tackles the ODGR problem from a novel perspective. It learns a \textbf{metric} for comparing the observation traces leading to different goals by utilizing deep metric learning. As done in previous methods that measure similarity between different time-series~\cite{Mueller_Thyagarajan_2016}, GRAML employs a Recurrent Neural Network (RNN) in a Siamese network architecture to \textbf{embed} sequences of observations. 
Its training aims that embeddings of sequences to the same goal would be closer to one another than embeddings of sequences to different goals. 
Upon receiving new goals, GRAML embeds any sequence to a new goal.  
These sequences can be received from a domain expert or be self-generated. 
 Then, upon receiving an input sequence, its embedding is measured against the embeddings of sequences of all goals, and the closest goal is returned. 

This work introduces several enhancements beyond existing GR:
First, GRAML offers a novel and robust model-free, self-supervised approach across continuous and discrete environments. It is aimed to improve recognition speed and applicability in complex domains, with little to no decrease in accuracy.
Second, the suggested framework addresses challenges unique to ODGR, posed by environments with changing goals, a facet often overlooked by previous GR work. 
%
Third, this study 
presents a set of ODGR benchmark environments to test its components. These environments were designed to represent as realistic and applicative scenarios as possible by generating suboptimal and diverse goal-directed observations using a variety of agents.


\section{Preliminaries}
We overview the definitions required for GR
and provide background on Metric Learning and LSTM networks. 
%


\subsection{Goal Recognition (GR)}
A GR problem consists of an \textit{actor} and an \textit{observer}, and it is articulated through the observer's perspective to infer the actor's goal. We use a common definition for goal recognition \cite{RamirezGeffner09,Pereira_Meneguzzi_2023}:

\begin{definition}
A \textbf{Goal Recognition problem} is a tuple $\langle T, G, O\rangle$, such that $T=\langle S,A \rangle$ is a domain theory where $S$ is the state space and $A$ the action space of the observed actor, $G$ is a set of goals, and $O$ is a sequence of observations, which are state-action tuples. 
The output of a GR problem is a goal $g \in G$ that best explains $O$.\
\end{definition}

\noindent Notice that this definition is very broad to enable different solution approaches. GR was often defined with $G \subseteq S$, but this work does not make this assumption regarding the nature of goals in the environment, and they can be any computable set as we see fit. For example $G$ can be equal to $S$, its power set, or a set of labels.
The definition assumes the actor is pursuing one goal exclusively, but 
there can be several different metrics to quantify the likelihood of one goal over another.
On the other hand, some GR approaches can handle observations that are either a sequence of actions or states but do not require both. 


In this work, a Domain Theory $T$ is modeled via an MDP:
\begin{definition}
\label{def:mdp}
    A Markov Decision Process $M$ is a tuple $\langle S, A, T, r, \gamma, \rho_0\rangle$, where $S, A, \gamma$
 and $\rho_0$ denote the state space, action space, discount factor, and distribution of initial states, respectively. $T:S\times A \times S \to [0,1]$ is the transition function, and $r: S \times A \to \mathbb{R}$ is the reward function.
\end{definition} In Reinforcement Learning (RL), the agent learns a stochastic \textbf{policy} $\pi:S \times A \rightarrow [0,1]$, that is aimed at maximizing the expected discounted cumulative return 
\begin{equation*}
\mathbb{E}_{a_t \sim \pi(\cdot \mid s_t)
, \, s_{t+1} \sim \mathcal{T}(\cdot \mid s_t, a_t)
} \left[ \sum_{t} \gamma^t r(s_t, a_t) \right]
\end{equation*}
This work focuses on infinite-horizon MDPs with a discount factor $0 < \gamma < 1$, where the state and action spaces can be either discrete or continuous. The RL tasks considered in this work are episodic, terminating upon task completion or after a maximum number of steps is reached.
Additionally, this work utilizes Goal-conditioned RL (GCRL)~\cite{liu2022goalconditioned}, an approach where an agent is trained towards multiple possible goals in an episode. A GCRL agent learns a \textbf{goal-conditioned policy} $\pi: G \times S \times A \rightarrow [0,1]$, thus exhibiting different behaviors for different goals.

\paragraph{Online Dynamic Goal Recognition}
Shamir et al.~\shortcite{shamir2024RLC} decompose the GR process into iterated phases with distinct inputs and outputs such that each phase can occur once or more, thus shifting from a GR problem to an Online Dynamic Goal Recognition (ODGR):

\begin{definition} \label{ODGR}
An ODGR problem is a tuple 
$\langle T, \langle G^i, \{O\}^i \rangle_{i\in 1..n}\rangle$, 
where $T = \langle S, A \rangle$ is a domain theory, 
$G^i$ is a set of goals, and $\{O\}^i$ represents a set of observations sequences, such that 
the $j$'th observation sequence given after the arrival of $G^i$ as part of $\{O\}^i$ is denoted as $O^i_j = \langle o_{j,1}^i, o_{j,2}^i \ldots \rangle = \langle \langle s_{j,1}^i, a_{j,1}^i \rangle, \langle s_{j,2}^i, a_{j,2}^i \rangle, \ldots \rangle$. 

\end{definition}

For each \( i \), the arrival of \( G^{i+1} \) replaces \( G^i \) as the active set of goals, where the goal sets may or may not overlap. For all $i, j$, an algorithm for solving ODGR is expected to return a goal $g \in G^i$ that best explains $O_j^i$ upon its arrival. The final output is a set of sets of goals: $G^* = \{\{g_1^1, g_2^1, ...\}, \{g_1^2, g_2^2, ...\}, ..., \{g_1^n, g_2^n, ...\}\}$ where $g_j^i$ is the most likely goal upon receiving $O_j^i$. 

For each domain, several \textit{instances} can be constructed at different times according to the set of potential goals $G^i$ the actor may be pursuing. For each instance, a single recognition \textit{problem} is an observation trace that needs to be explained. We can split an ODGR problem into several time intervals, depending on the reception of each input type, $T$, $G$, and $O$. Within an ODGR problem, one GR domain exists: as many GR instances as the number of times the goals changed and as many GR problems as the number of observations given as a query to the framework.
We name the time intervals according to the reception of each input $T$, $G$, $O$: 

\begin{figure}[t]
  \centering
  \includegraphics[width=0.48\textwidth]{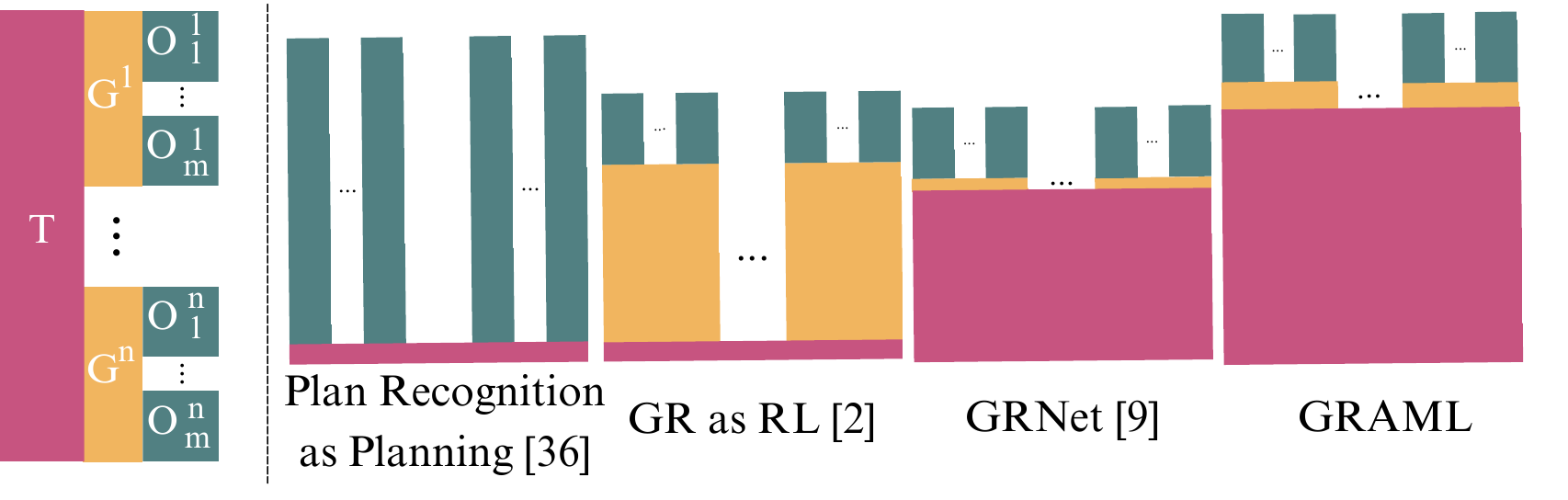}
  \caption{Depiction of ODGR intervals and the inputs at each (left) and a rough approximation of the time spent on each phase by representative GR frameworks (right). } 
  \label{fig:time_spent}
\end{figure}
\begin{description}
    \item [T] \textbf{Domain Learning Time} is the duration from receiving the domain theory $T$ until concluding the domain-specific processing and being prepared to receive $G$.
    \item [G] \textbf{Goals Adaptation Time} is the duration from receiving $G$ until completing the inner-state changes and becoming ready to perform inference.
    \item [O] \textbf{Inference Time} is the time from getting an observation sequence $O$ until the algorithm outputs the goal.
\end{description}


An ODGR framework can receive a new piece of input before completing the processing of the previous one. 
These definitions remain independent of the time steps assigned to each input. However, for simplicity, in this work, we assume that inputs must \textit{precede} one another. 
Figure \ref{fig:time_spent} (left) illustrates the connection between the different phases in an ODGR problem. 
Most of the computation time in symbolic GR approaches occurs at inference time and often depends on the number of goals $|G|$ \cite{ijcai2021p616}. 
Recent learning-based approaches require less time during the inference phase by pre-processing the domain theory. However, they either necessitate pre-processing the domain for every newly emerging goal \cite{amado2022goal} or result in longer domain learning times \cite{chiari2022goal}. GRAML aims to reduce the goal adaptation time while maintaining a short inference time at the expense of extended domain learning time.
%
A comparison of these approaches is shown in Figure \ref{fig:time_spent} (right).

\begin{figure*}[ht]
  \centering
  \includegraphics[trim=1.3cm 9cm 1.1cm 9cm,clip,width=\textwidth]{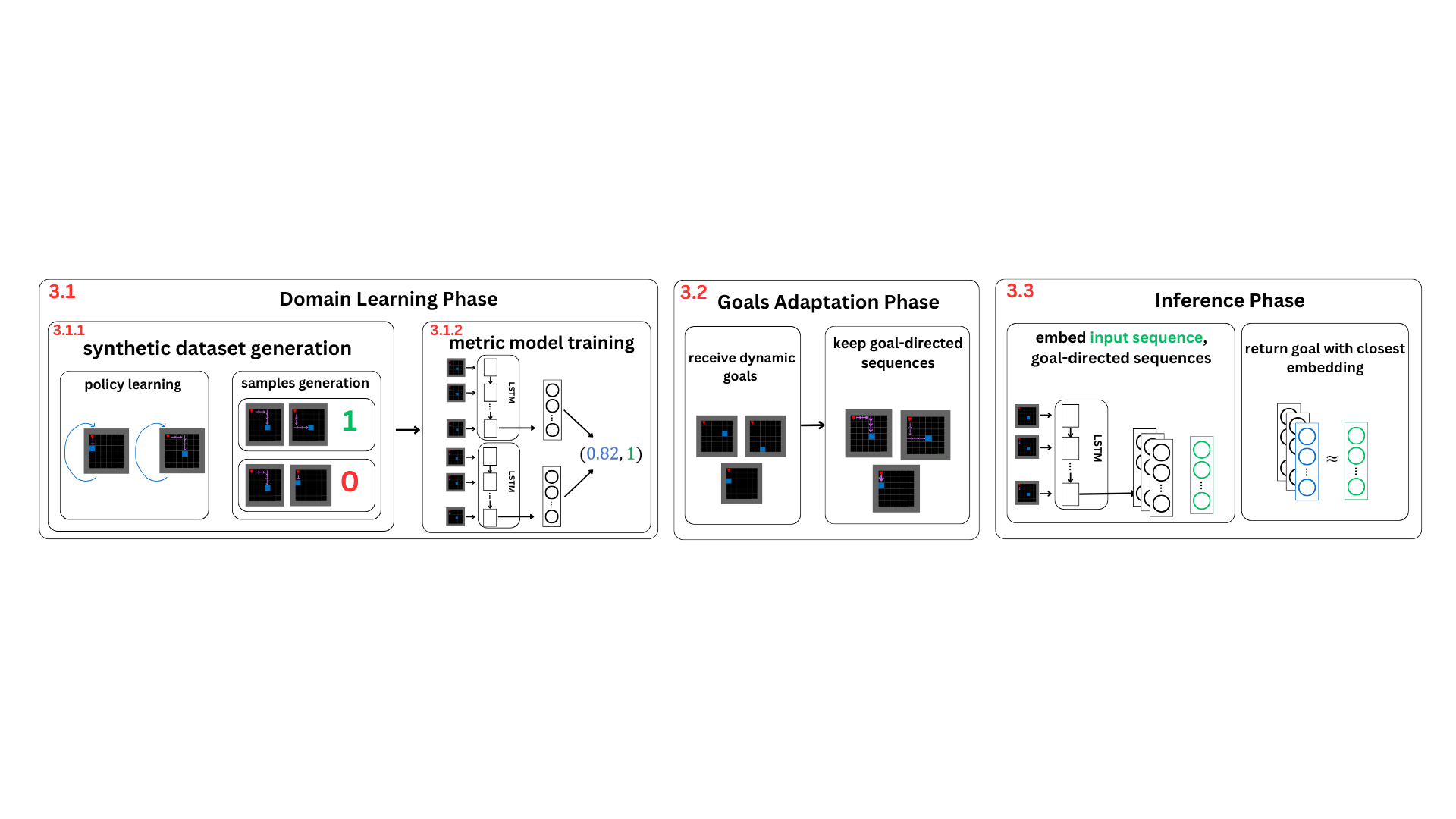}
  \caption{GRAML's implementation for the different phases of an ODGR problem. The numbers at the top left of each box reflect the section in which each component is discussed.}\label{fig:GRAML}
\end{figure*}

\subsection{Metric Learning}

The primary purpose of metric learning is to learn a new metric where the distances between samples of the same class are expected to be small, and distances between samples of different classes are expected to be larger. This work aims to learn such a metric where sequences of observations from distinct goal-oriented agents are distant from each other but close if the sequences lead to the same goal. 
Similarity metrics that do not use Deep Neural Networks have limited capabilities due to their inability to represent non-linear structures, such as state-action observation sequences, and thus they cannot be utilized for GRAML \cite{duanIEEE2018,duan2020IEEE,zheng2019hardnessaware}. 
%
%
One common approach for representing such observation sequences for different possible goals (a multivariate time series) is Long Short-Term Memory (LSTM) for modeling time series similarities \cite{Mueller_Thyagarajan_2016,Rakthanmanon2012SearchingAM}. In these models, the last hidden layer represents the input sequence, and the L1 or L2 distance between the representations is the global distance. 
It has been shown that when using DNNs, a metric learned in a set of classes can be generalized to new classes. This work builds upon these generalization abilities when new goals are introduced.

\section{Goal Recognition as Metric Learning}



GRAML employs a metric-learning model trained on sequences that converge to prototype goals, forming a representation in which observation sequences from agents with distinct goal orientations are widely separated while sequences converging on the same goal are closely clustered. 
Figure \ref{fig:GRAML} overviews the different phases of GRAML: During the domain learning phase (Section \ref{sec:learn}), 
GRAML undergoes self-supervised training using agents trained for the goal recognition domain to generate labeled samples. In the goal adaptation phase (Section \ref{sec:adapt}), GRAML requires sequences for each newly introduced goal. Lastly, during the inference phase (Section \ref{sec:infer}), when presented with a new trace and a GR problem, GRAML embeds the goal-directed traces it got to create mappings from each dynamic goal’s sequence into the embedding space, capitalizing on the embedding network’s generalization capabilities acquired during the previous phase. GRAML identifies the goal whose embedding is closest to the input's trace embedding.

We adopt two distinct strategies for GRAML: the \textit{base-goals} approach, referenced to as \textbf{BG-GRAML} and the \textit{goal-conditioned RL} approach, referenced to as \textbf{GC-GRAML}. As the names imply, GC-GRAML utilizes a single GCRL agent that learns a goal-conditioned policy during domain learning and uses the same policy to generate traces in the goal adaptation phase. In contrast, BG-GRAML involves training separate goal-directed agents and using expert-generated samples or planners in the goal adaptation phase.  In the following sections, we outline each framework phase and describe how GRAML operates under both strategies.

\subsection{Domain Learning Phase}
\label{sec:learn}
In the domain learning phase, GRAML receives a domain theory $T = \langle S, A \rangle$ and performs a self-supervised training process split into two: dataset generation and model training. 


\subsubsection{3.1.1 Dataset Generation}
The dataset generation involves training agents on selected goals in the environment and stochastically generating partial sequences based on the agents' policies and labeling each pair of sequences according to whether they lead to the same goal. Each sample consists of two sequences labeled as either 0 or 1, depending on whether they lead to the same goal.

In BG-GRAML, we first select $m$ base goals $\bar{G} = \langle \bar{g}^1, ..., \bar{g}^m \rangle$ that ideally cover the potential goal space. These goals are not a part of the ODGR problem formulation but a distinct set chosen by GRAML prior to receiving the first set of possible goals for recognition, $G^0$. We train individual goal-directed RL agents for each $g \in \bar{G}$, obtaining a set of policies $\bar{\Pi} = \langle \bar{\pi}^1, ..., \bar{\pi}^m \rangle$. 
Then, observation sequences are generated by randomly pairing agents trained towards goals $\bar{g}^i, \bar{g}^j \in \bar{G}$ and have them produce traces ${O_i}, {O_j}$ that end at the assigned goal, creating a labeled sample $\langle {O_i}, {O_j}, y \rangle$. If $i$ equals $j$, 
then $y$ equals $1$. Otherwise, $y$ equals $0$, since these traces lead to separate goals. For more robust training, parts of the sequences are randomly removed, creating both consecutive and non-consecutive traces at varying lengths to associate all kinds of traces that might be encountered in the inference phase with the same goal. Section \ref{sec:eval} details the generation of these input variations as used in our empirical work.

GC-GRAML follows a similar process, buts instead of multiple agents, a single GCRL agent 
is trained on $\bar{G}$, which can also be defined as a continuum or discrete set. The sample generation is also done similarly, but in this case, a single goal-conditioned policy generates the traces. 
Another important distinction is that in GC-GRAML, the GCRL agent, along with $\bar{\pi}^*$, is kept to be used in the goal adaptation phase.
\subsubsection{3.1.2 Model Training}
 Given an existing dataset of $n$ labeled samples $D = \{\langle {O^1}, {O^2}, y \rangle_1, ..., \langle {O^1}, {O^2}, y \rangle_n\}$, the model training involves training a metric model $M$, structured as a Siamese network and comprised of 2 LSTM branches with shared weights. The input to each LSTM unit has a size of $s + a$, where $s$ and $a$ represent the sizes of the state and action representations in $T$, respectively. For every sample $\langle {O^1}, {O^2}, y \rangle_i$, the model processes ${O^1}, {O^2}$ through branches 1 and 2 to obtain 2 embeddings $v^1, v^2 \in \mathbb{R}^k$. The model then calculates an exponential transformation on the Manhattan distance metric between $v^1, v^2$, to produce a similarity score: \begin{equation}
 \label{similarity} 
 \hat{y} = \text{similarity}(v^1, v^2) = e^{-\sum_{i=1}^k \left| v^1[i] - v^2[i] \right|} 
 \end{equation} where brackets denote specific elements of the vector. The more similar the numerical values are in each corresponding element of the embeddings, the closer the similarity score is to 1. 
 The model is penalized using the binary cross entropy loss function:
 \begin{equation}
    L = -\left( y \log(\hat{y}) + (1 - y) \log(1 - \hat{y}) \right)
\end{equation}

\subsection{Goal Adaptation Phase}
\label{sec:adapt}
The goal adaptation phase starts upon receiving the $i$-th set of active goals, $G^i=\{g_1, ..., g_{|G^i|}\}$, The main challenge GRAML faces is obtaining a set of libraries of goal-directed sequences $O^* = \langle \{O^1, ..., O^s\}_{g_1}, ..., \{O^1,..., O^s\}_{g_{|G|}}\rangle$, where $s$ denotes the library size and is a hyperparameter that can be tuned. 
%
If a domain expert can provide $O^*$, the adaptation is instant (discarding the domain expert's efforts). If such an expert is unavailable, there are two alternatives corresponding to the strategy by which GRAML operates:
In BG-GRAML, an off-the-shelf planner that operates in MDPs, such as MCTS, is leveraged to produce $O^*$, including one or more goal-directed sequences for each $g \in G^i$.
In GC-GRAML, the goal-conditioned policy $\bar{\pi}^*$ generates the sequences, using stochastic action predictions to achieve diverse sequences, counting on its generalization abilities. 

\subsection{Inference Phase}
\label{sec:infer}
 This phase starts with an active set of goals $G^i$, and a single observation $O_j^i = \langle o_{j,1}^i, o_{j,2}^i \ldots \rangle.$ The objective of this phase is to return the goal  $g \in G^i$ to which $O_j^i$ most likely leads. Using the LSTM from the learning phase and the sequences for the active set of goals, $O^*$, the goal whose embedding is closest to the embedding of the input sequence $O$ is returned.

 Using $O^*$, GRAML produces embeddings libraries $\langle\{v^1,...,v^s\}_{g_1}, ..., \{v^1,...,v^s\}_{g_{|G^i|}}\rangle$ corresponding to every goal-directed library $\{O^1, ..., O^s\} \in O^*$ . These embeddings could potentially be generated during the goal adaption phase to decrease online inference runtime. However, truncating the full sequences $O^*$ to $O^i_j$'s length after receiving showed increased accuracy in our experiments.
 It then produces yet another embedding $v$ for the input sequence $O_j$, and returns the goal that maximizes the average similarity with all sequence embeddings for a specific goal with $v$:
\begin{equation}
    g = \arg \max_{g \in G^i} \, \frac{1}{s} \sum_{j=1}^s \text{similarity}(v_{g}^j, v)
\end{equation} Where $v_g^j$ denotes embedding number $j$ in the library of a goal $g$.
This averaging method improves recognition accuracy by accounting for different trajectories for the same goal. However, it introduces a trade-off: as the value of $s$ increases, the runtime also increases. Constructing larger libraries is likely to improve GRAML’s accuracy, but as demonstrated in the empirical evaluation, the library generation time grows linearly with $s$ for both BG-GRAML and GC-GRAML.

\section{Evaluation}
\label{sec:eval}
We implemented GRAML using Python.\footnote{Link to the repository will be added post-acceptance.} All experiments were conducted on a commodity Intel i-7 pro.



\paragraph{Variables}
We vary our experiments across the following dimensions to test GRAML under different settings:
\begin{itemize}
\item \textbf{Discrete or Continuous state/action environments}. In our experiments, we have two environments with discrete state and action spaces (Minigrid-SimpleCrossing and Minigrid LavaCrossing) and four environments with continuous ones (Parking, PointMaze-Obstacle,
PointMaze-FourRooms and Panda-Gym Reach). 

\item \textbf{Consecutive and non-consecutive partial observations}. A \textit{consecutive} sequence is a partial sequence of observations from a starting state to the goal, where the missing parts are from a certain observation until the goal. 
A \textit{non-consecutive} sequence is a partial sequence of observations from a starting state to the goal, where the missing parts are randomly chosen from the sequence. 
For each environment, we tested observation sequences that are $30\%, 50\%, 70\%$, and $100\%$ of the full sequence, both consecutively and non-consecutively.

\item \textbf{Suboptimal Observations}. 
At inference, a stochastic policy generates sub-optimal observation sequences.

\item \textbf{Initial States}. The initial steps of the model are taken randomly to foster the creation of diverse plans, similar to GRAML's dataset generation process. 

\end{itemize}

For each ODGR problem generated, there is a single goal adaptation phase followed by several inference phases where each goal in the active set is considered the true goal with every observability configuration. 
This process yields $5$ (average number of active goals $|G^0|) \times 8$ (observability $\%$ and consecutive/non-consecutive) $=40$ unique inference phases. Since there are five separate ODGR problems for every environment, there are five goal adaptation phases and 200 GR problems for every one of the six environments.

\begin{figure}
  \centering
  \includegraphics[width=0.45\textwidth]{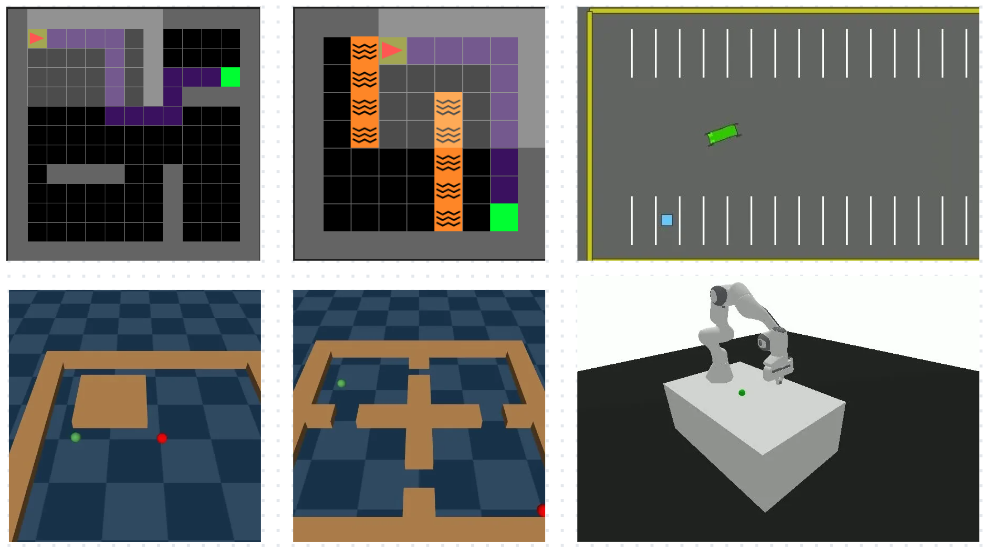}
  \caption{The evaluation environments from top-left to bottom-right: Minigrid-SimpleCrossing, Minigrid-LavaCrossing, Parking, PointMaze-Obstacle, PointMaze-FourRooms and Panda-Gym.}
  \label{fig:domains}
\end{figure}

\subsubsection{Data Sets}
Building on the GCRL survey and the benchmark environments suggested at ApexRL \footnote{\url{https://github.com/apexrl/}}, we form a collection of GR problems from several sets of environments that adhere to the Gymnasium API \footnote{\url{https://gymnasium.farama.org/}}, with detailed descriptions of each in Appendix \ref{app:envs}. We consider two custom Minigrid environments, two custom PointMaze environments \cite{fu2020d4rl} 
from the Gymnasium-Robotics package, the Parking environment from the highway-env package \cite{highway-env}, and the Reach environment from PandaGym \cite{gallouedec2021pandagym}. 
%
Figure \ref{fig:domains} showcases these environments.

\paragraph{Algorithms}
Our primary evaluation is of the BG-GRAML and GC-GRAML implementations. During the goal adaptation phase of BG-GRAML, a self-implementation of MCTS was utilized to generate plans for embedding in the discrete case.  In the continuous scenarios, when using expert examples, the algorithm used to produce them differed from the one used by the input generation in the inference phase. 

We compare BG-GRAML and GC-GRAML to the GRAQL \cite{amado2022goal} algorithm in discrete domains and DRACO \cite{nageris2024goal} in continuous domains. 
These algorithms engage in complete policy learning for each dynamic goal in the goal adaptation phase, resulting in longer runtime. 
Another variation, named GC-DRACO, learns a goal-conditioned policy once during the learning phase. It is used in our experiments to compare against GC-GRAML.
For agent training, we use Deep RL algorithms from Stable-Baselines3 \cite{stable-baselines3} in the continuous environments and apply the tabular Q-Learning originally used by GRAQL for discrete environments throughout the framework's phases.
Beyond varying the environments' properties, experimental diversity is further achieved by ensuring that the RL algorithms used to generate sequences for each phase are distinct and rotate between PPO, SAC, and TD3. 


Next, we showcase how the learned embeddings enable us to measure the distance between goals (Figure \ref{fig:conf_mat}). We continue to evaluate the trade-off between BG-GRAML and GC-GRAML (Figure \ref{fig:strategies}). Then, we showcase GRAML's performance in inference accuracy given varying observability (Table \ref{tab:accuracies_table}) and discuss its runtime. Lastly, we examine the effect the number of base goals $\bar{G}$ and the number of goals in the active goal set $G^i$ has on the recognition accuracy (Figure \ref{fig:increasing_goals}).

\subsection{Embedding Similarity}

\begin{figure}[t]
  \centering
  \includegraphics[trim=4.2cm 1cm 1cm 1cm, clip, width=0.485\textwidth]{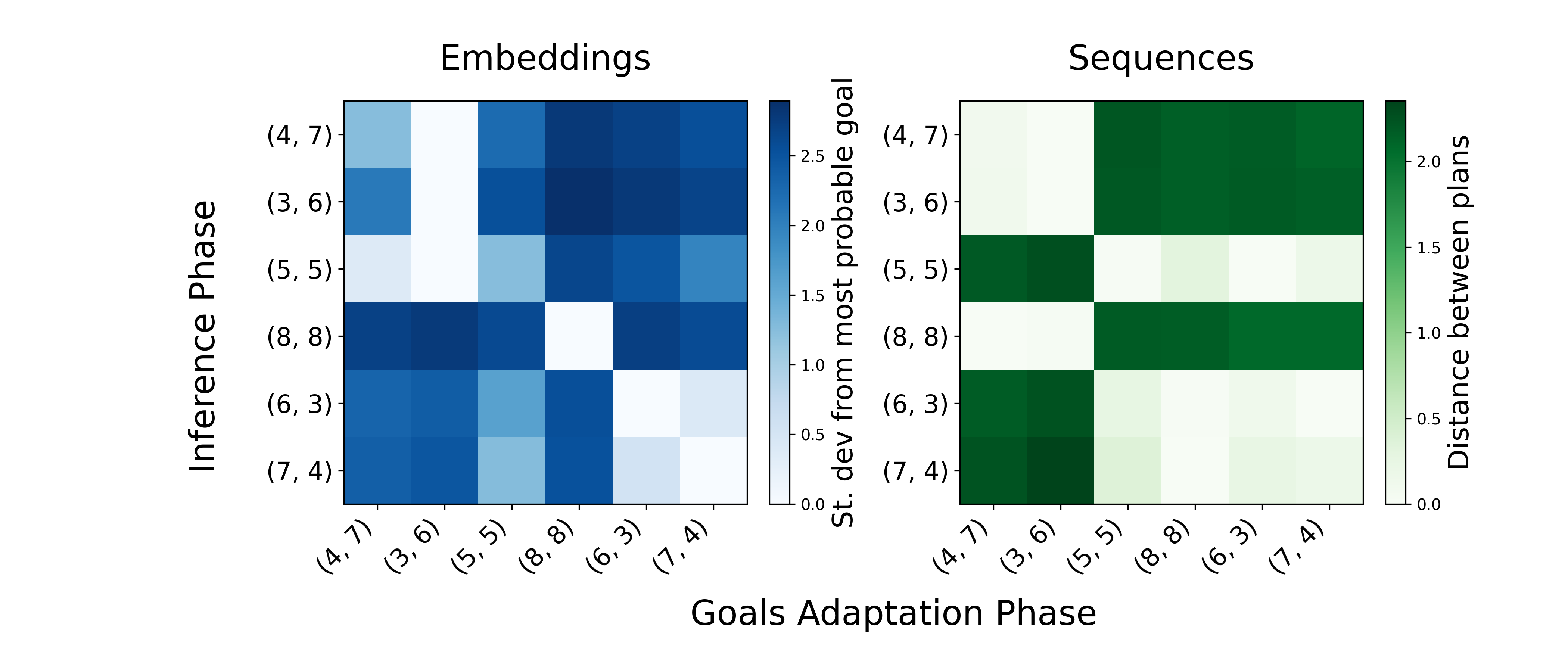}
  \caption{Confusion matrices for plan similarity and recognition confidence in the inference phase for GRAML across multiple tasks in the PointMaze environment where the goals differ from the set of base goals from the domain learning phase.}
  \label{fig:conf_mat}
\end{figure}

Before evaluating the performance of GRAML for GR, we verify that the learned model embeds sequences leading to the same goal closer together. 
This is especially important when execution can vary, and different distinct sequences can lead to the same goal. Figure \ref{fig:conf_mat} shows that BG-GRAML
is able to create an embedding in which different sequences to the same goal get closer embeddings. The right matrix shows similarity scores between sequences, where cell $(i, j)$ represents the similarity between the sequence reaching goal $i$, which was received as input to the inference phase, and the sequence leading to goal $j$ that was received in the latest goal adaptation phase. The left matrix follows a similar structure where, instead of sequence similarities, it shows the similarities between the embeddings achieved for those sequences.
For example, despite the lack of similarity between the sequences leading to the goal (8,8) in the goal adaptation phase (Figure \ref{fig:conf_mat} (right)), GRAML (Figure \ref{fig:conf_mat} (left)) learns an embedding in which these sequences are close to one another.



\subsection{GRAML Strategies}
\begin{figure}
    \centering
    \includegraphics[trim=1.5cm 0cm 2cm 1.2cm, clip, width=\linewidth]{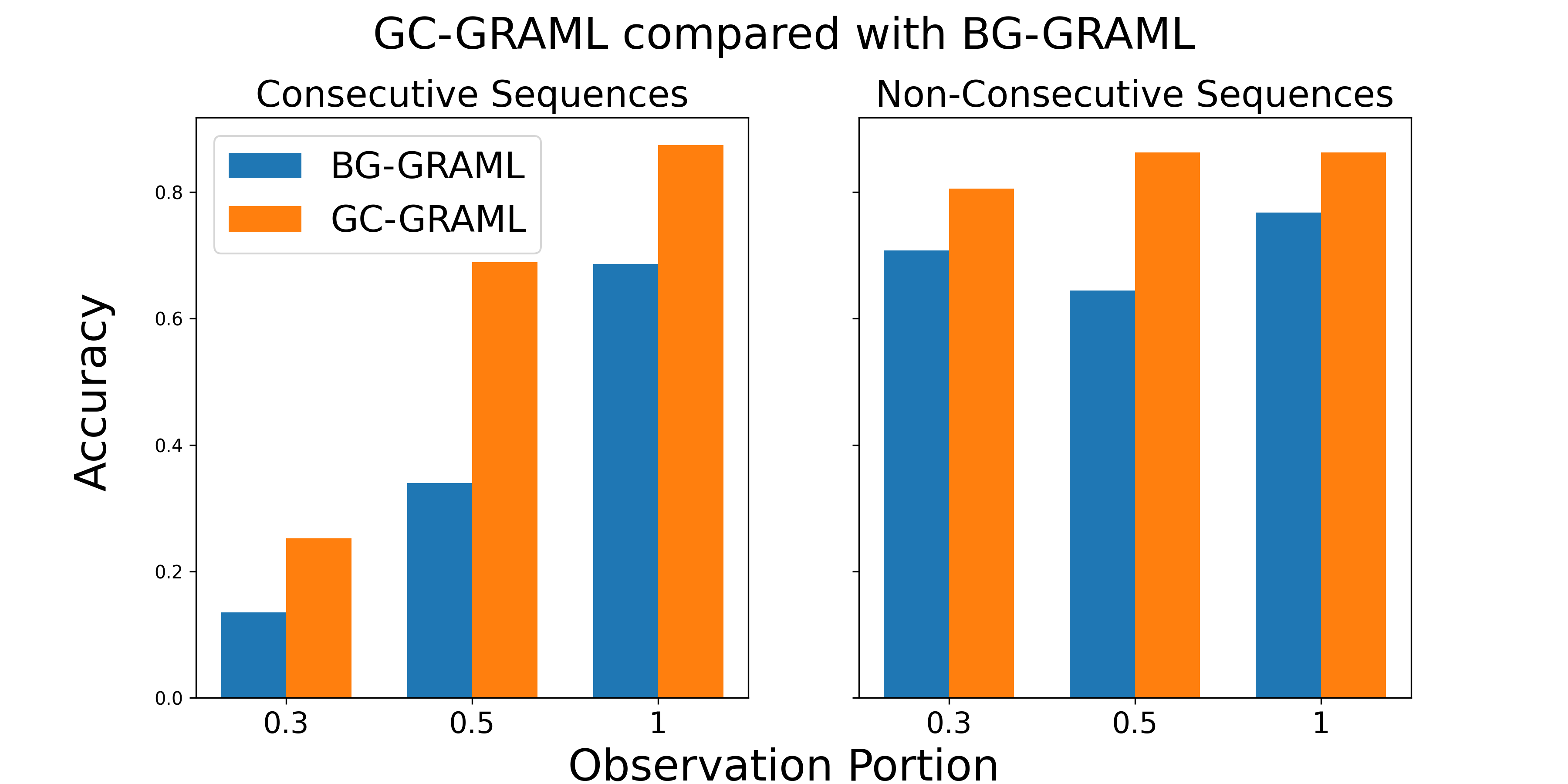}
    \caption{BG-GRAML and GC-GRAML accuracy in Parking.}
    \label{fig:strategies}
\end{figure}

Next, we compare BG-GRAML and GC-GRAML's accuracy given sequences of changing observability when we aim to adapt quickly to new goals while minimizing
the dependence on a domain expert. 
To do so, both BG-GRAML and GC-GRAML are fed with a single sequence per goal in $(s=1)$. Each single-goal agent was trained for $300,000$ timesteps, and the goal-conditioned agent was trained for $1$ million timesteps.
As GC-GRAML utilizes the goal-conditioned policy $\pi^*$, the number of base goals it can use does not affect domain learning time. Hence, the size of $\bar{G}$ was set to 20. Contrarily, in BG-GRAML, this value was set to $5$. Figure \ref{fig:strategies} shows that GC-GRAML benefits from the variety of sequences the model observed during learning and can consistently generalize and recognize observation sequences it was not trained on better than BG-GRAML.

\subsection{Inference Performance}
Given a GR problem, GRAML's performance is compared with the chosen GR as RL algorithms. 
In each of the scenarios presented in Section \ref{sec:eval}, a set of base goals $\bar{G}$ was chosen, along with sets of 3-9 active goals $G^0$ for the sole goal adaptation phase in the experiment. Every $g \in G^0$ was once chosen as the true goal, yielding a set of 200 GR problems per scenario on average, depending on the size of $G^0$.

\paragraph{GR accuracy}

\begin{table*}[ht]
    \caption{Accuracy for Different Domains, Sequence Lengths, and Algorithms, averaged over 200 instances (with std) for configurations varying by observation type and observability $\%$. Minigrid is further averaged on Minigrid-SimpleCrossing and Minigrid-LavaCrossing, and PointMaze on PointMaze-Obstacle and PointMaze-FourRooms. BG-G is BG-GRAML, GC-G is GC-GRAML, and GC-D is GC-DRACO.}
    \label{tab:accuracies_table}
    \centering
    \begin{adjustbox}{width=\textwidth}
        \begin{tabular}{l*{14}{c}}
            \toprule
            \multirow{2}{*}{Env.} 
            & \multicolumn{6}{c}{Consecutive} 
            & \multicolumn{8}{c}{Non-consecutive} \\ 
            \cmidrule(lr){2-7} \cmidrule(lr){8-15}
            & \multicolumn{2}{c}{30\%} & \multicolumn{2}{c}{50\%} & \multicolumn{2}{c}{70\%} 
            & \multicolumn{2}{c}{30\%} & \multicolumn{2}{c}{50\%} & \multicolumn{2}{c}{70\%} & \multicolumn{2}{c}{100\%} \\ 
            \cmidrule(lr){2-3} \cmidrule(lr){4-5} \cmidrule(lr){6-7} \cmidrule(lr){8-9} \cmidrule(lr){10-11} \cmidrule(lr){12-13} \cmidrule(lr){14-15}
            & BG-G & GRAQL & BG-G & GRAQL & BG-G & GRAQL
            & BG-G & GRAQL & BG-G & GRAQL & BG-G & GRAQL & BG-G & GRAQL \\
            \midrule
            Minigrid
            & 0.35(0.18) & \textbf{0.40}(0.19) & 0.44(0.20)  & \textbf{0.51}(0.22) & 0.55(0.19) & \textbf{0.58}(0.21)
            & 0.63(0.24) & \textbf{0.66}(0.22) & 0.77(0.17) & \textbf{0.81}(0.13) & \textbf{0.83}(0.17) & 0.84(0.20) & \textbf{0.82}(0.18) & 0.90(0.15) \\
            \midrule
            & BG-G & DRACO & BG-G & DRACO & BG-G & DRACO
            & BG-G & DRACO & BG-G & DRACO & BG-G & DRACO & BG-G & DRACO \\
            \cmidrule(lr){2-15}
            PointMaze
            & \textbf{0.51}(0.20) & 0.50(0.23) & 0.63(0.21) & \textbf{0.71}(0.20) & \textbf{0.83}(0.17) & 0.82(0.18)
            & 0.75(0.30) & \textbf{0.81}(0.21) & 0.79(0.29) & \textbf{0.84}(0.20) & 0.78(0.27) & \textbf{0.84}(0.16) & \textbf{0.80}(0.24) & 0.84(0.16) \\
            \midrule
            & GC-G & GC-D & GC-G & GC-D & GC-G & GC-D
            & GC-G & GC-D & GC-G & GC-D & GC-G & GC-D & GC-G & GC-D \\
            \cmidrule(lr){2-15}
            Parking
            & \textbf{0.43}(0.26) & 0.42(0.22) & 0.45(0.24) & \textbf{0.49}(0.20) & \textbf{0.76}(0.20) & 0.54(0.19)
            & \textbf{0.84}(0.16) & 0.55(0.19) & \textbf{0.88}(0.19) & 0.56(0.19) & \textbf{0.88}(0.17) & 0.66(0.18) & \textbf{0.89}(0.16) & 0.71(0.18) \\
            Panda
            & \textbf{0.45}(0.23) & 0.35(0.16) & \textbf{0.74}(0.17) & 0.56(0.20) & \textbf{0.82}(0.16) & 0.79(0.03)
            & \textbf{0.83}(0.20) & 0.82(0.20) & \textbf{0.91}(0.15) & 0.90(0.11) & 0.92(0.13) & \textbf{0.94}(0.10) & 0.92(0.14) & \textbf{0.95}(0.08) \\
            \bottomrule
        \end{tabular}
    \end{adjustbox}
\end{table*}

Table \ref{tab:accuracies_table} summarizes the execution results in terms of accuracy. Notice that due to the ambiguity of traces (especially with consecutive observations and low observability), even a perfect recognizer might not be able to reach perfect recognition.
The performance of the GRAML variants consistently demonstrated strong results, comparing to or surpassing the GR as RL versions in many scenarios despite its decreased adaptation time. Notably, GRAML benefits from increased observability levels compared to algorithms that were trained directly on the current goal set $G^0$.


\paragraph{GR runtime}
Recall that GRAML aims to optimize goal adaptation and inference times. We compare the computation time during goal adaptation in GRAML to the goal learning in GRAQL and DRACO, as they need to learn these goals for every GR instance. We compared all supported domains, averaging results over 200 instances and selecting the best-performing variant from each algorithm. Recall that G and O denote goal adaptation and inference times, respectively. Algorithms exceeding two hours for goal adaptation were deemed to have timed out. In the Minigrid domain, GRAQL timed out during G, with O averaging 0.04 seconds, while BG-GRAML incurred $G=0$ due to the absence of computations and achieved an average O of 0.03 seconds. In the PointMaze domain, DRACO timed out during G and had an average O of 2.56 seconds, whereas BG-GRAML incurred $G=0$ and achieved an average O of 1.47 seconds. In the Parking domain, GC-DRACO timed out during G and averaged $O=2.49$ seconds, while GC-GRAML required 36 seconds for G and averaged $O=2.17$ seconds. In the Panda domain, GC-DRACO timed out during G and averaged $O=2.24$ seconds, while GC-GRAML required 87 seconds for G and 5.29 seconds for O. BG-GRAML achieves $G=0$ by leveraging expert sequences. For reference, in the Parking domain with $|G_i|=5$, DRACO required ten hours for goal adaptation. 

Additional times are the acquisition of the base goals sequences during the self-supervised learning and potential plan time in the goals adaptation of BG-GRAML. Running BG-GRAML with MCTS in discrete environments for planning instead of receiving a sample from an expert yields a planning time of 75 seconds per goal on average. Considering the time it takes to train agents towards every active goal as the timeout, creating a library of sequences per goal in a problem with five goals on average takes 18 minutes, about 4 times faster than training agents for every goal in the active set.

\subsection{Goal Set Size}
\begin{figure}
    \centering
    \includegraphics[trim=0.15cm 0.1cm 0.1cm 0, clip, width=\linewidth]{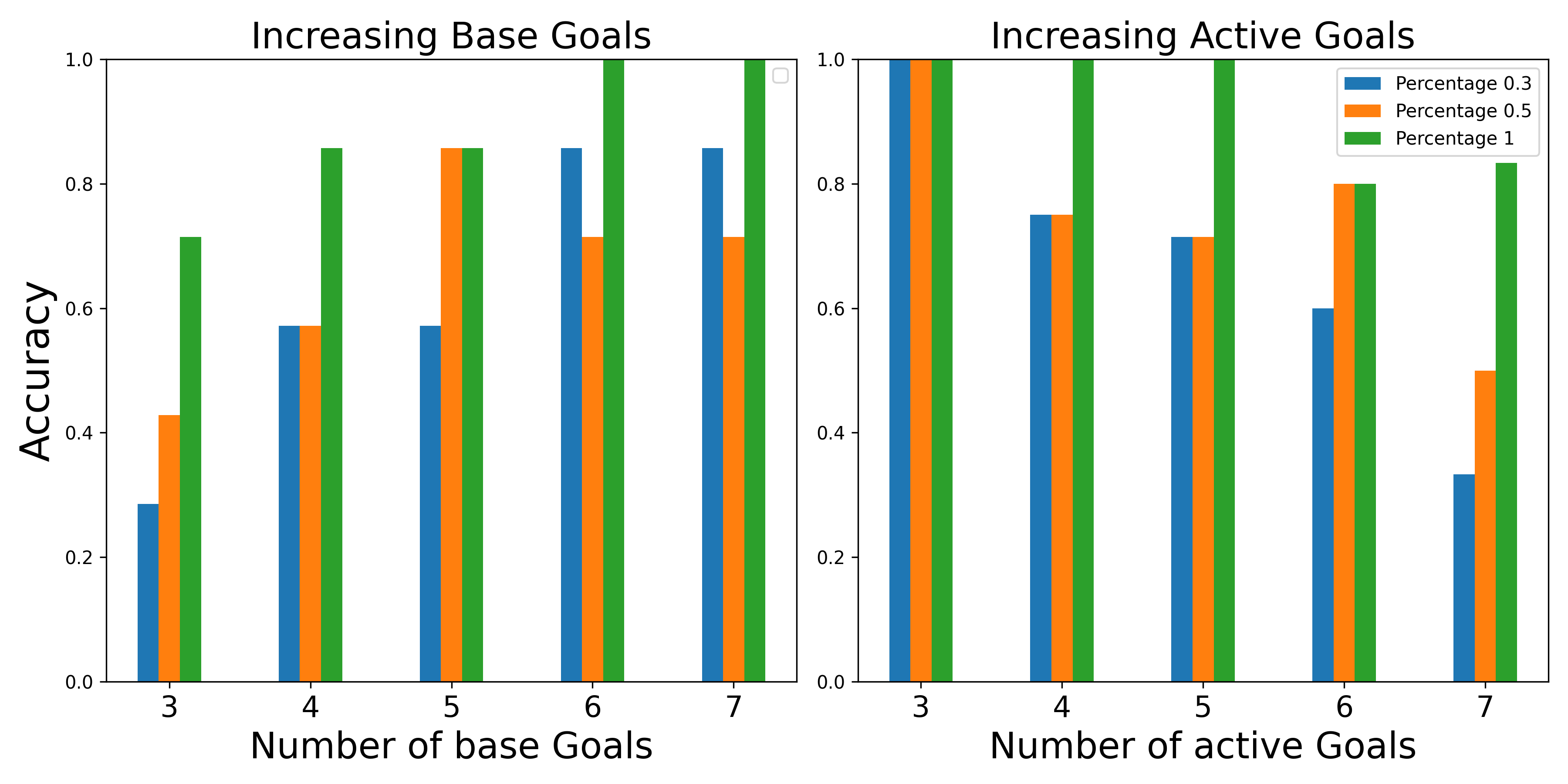}
    \caption{BG-GRAML's accuracy as a factor of the number of newly introduced goals.}
    \label{fig:increasing_goals}
\end{figure}


Lastly, we report the influence of the goal set size ($|\bar{G}|$) for the learning phase of GRAML and the active goal set size ($|G_i|$) on BG-GRAML's performance in the Parking environment. 
Figure \ref{fig:increasing_goals} (left) shows that an increasing number of base goals improves the algorithm's accuracy during recognition. This improvement is due to the wider variety of goal sequences to train on, as well as the sheer number of diverse goals, which, in turn, leads to higher generalization abilities.
Figure \ref{fig:increasing_goals} (right) shows an opposite trend as the number of active goals increases. This is unsurprising; as more goals are introduced, the possible similarity between sequences leading to them increases, making the goals harder to disambiguate.

There are a couple of instances where, on average, lower observability leads to better recognition than higher observability. We attribute this behavior to the random way the sequences were generated for each experiment (recall that the policy used for this generation process is stochastic). Still, the results show a mostly consistent behavior.



\section{Related Work}
This section provides an overview of recent work, emphasizing the strengths and weaknesses of each methodology under different assumptions regarding input timing and the potential changes in goals presented by ODGR problems. We split existing work into Model-Based GR (MBGR) and Model-Free GR (MFGR) ~\cite{geffner2018model}:


\paragraph{Model-Based GR (MBGR)}
In MBGR, the recognition process entails utilizing a pre-defined and clearly stated model that encompasses the characteristics of an environment along with the actions that can be executed within that environment~\cite{ijcai2021p616,mirsky2021symbolic}.
Traditional MBGR usually exploits planning and parsing techniques \cite{baker2009planning,geib2009probabilistic,RamirezGeffner09,ramirez2011goal}. 
%
%
Although the domain learning for these algorithms is non-existent in terms of compute, it requires a domain expert to manually construct these models.



\paragraph{Model-Free GR (MFGR)}
In MFGR, the recognizer has no access to the underlying model that describes the properties and dynamics of the environment~\cite{geffner2018model}. 
Some approaches learn the model and then employ MBGR~\cite{asai2022classical,geib2018learning,su2023fast}, while others perform end-to-end model-free GR without explicitly learning the model~\cite{amado2022goal,borrajo2020goal,chiari2022goal,fang2023real,min2014deep}.
%
These approaches assume the presence of various inputs during training, and most do not explicitly address goal generalization. GRAML belongs to the MFGR family. Yet, unlike most MFGR methods, it alleviates the need for a dataset of solved GR tasks using self-supervised sequence generation.

The closest approach to GRAML, in terms of the underlying model, is GR as RL \cite{amado2022goal} that also uses an MDP representation of the actor's states and actions. This approach trains a policy for every potential goal. This computation is costly and cannot directly address ODGR. 
%
A further limitation of existing GR methods is their inherent reliance on a relatively narrow behavior considered ``optimal''. Almost all of the above GR approaches (excluding a few who address this issue implicitly \cite{chiari2022goal} or explicitly \cite{sohrabi2016plan}) use a single plan or policy which might incorrectly classify an observation as unrelated to a specific goal when it actually is, only via a new, diverse plan. 

While most approaches don't specifically tackle ODGR, Chiari et al. \shortcite{chiari2022goal} proposed a framework termed \textbf{GRNet} that generalizes to any set of goals without additional learning, suggesting a model-free solution suitable for dynamic environments. 
GRnet employs RNNs to handle observation sequences and assumes a dedicated learning phase on a given dataset of solved GR problems, wherein the domain theory is provided before any goals or observations, resulting in an extensive domain learning time. 
While this approach is promising, as with other planning-based methods, its reliance on fluent enumeration is critical for success, and while adaptation to continuous domains is conceivable with techniques like discretization, this area necessitates further research. As a comparison, GRAML handles continuous environments by having the LSTM output layer operate in an embedding space rather than in the state space.


\section{Conclusions and Future Work} 
\label{sec:conclusions}


This paper introduces GRAML, a new algorithm for GR and ODGR using Metric Learning. GRAML was designed to handle both discrete and continuous domains, prioritizing the \textbf{speed} of adaptation to emerging goals and the promptness of inference upon receiving a sequence of observations $O$.
Aiming at robust applicability, GRAML also targets accurate recognition of \textbf{suboptimal} and \textbf{highly-dissimilar} observation sequences that lead to a certain goal.
We achieve diversity across (1) input sequences to the same goals, used to distinguish the acting agent and the observing agents' underlying policies, and (2) inference phase and goal adaptation phase sequences, used to generate more challenging and non-trivial recognition of new goals for GRAML. 

This paper further analyzes how different variables of an environment, such as the domain characteristics, the variability in sequence generation, and the observation sequence format, can influence learning and inference performance for ODGR algorithms and GRAML in particular. These domain-agnostic attributes are crucial when evaluating an ODGR framework, and researchers are welcome to build upon this methodology when evaluating new ODGR approaches.

We propose and discuss two variations of GRAML: 
BG-GRAML, though potentially time-consuming due to its reliance on expert samples or planners, avoids dependence on the generalization abilities of a GCRL agent. In contrast, when it is possible to implement, GC-GRAML does not require additional information during the goal adaptation phase. It also
benefits from generalization by offering a more varied dataset from which the LSTM can learn. 

There are still many optimizations and improvements that could be incorporated into GRAML:
First, this paper did not discuss how base goals are chosen for learning, which may be crucial for the model's success. 
If an area of the state space is not covered via observation sequences during the domain learning phase, 
the model may be unable to recognize goals in that area correctly. 
In our results, training a GCRL agent with as many goals as possible was the most promising approach to solving this problem. It also prompts for optimizations at goal adaptation time, like the averaging method over a library of samples. Other methods can include eliminating outliers and clustering. 
Lastly, results show how the number of base goals impacts recognition, where base goals were manually selected for BG-GRAML and randomly for GC-GRAML. Future work could explore how to optimize not only the quantity but also the quality of base goals.




\bibliographystyle{named}
\bibliography{aaai24}

\clearpage

\appendix
\section{Environment Descriptions}
\label{app:envs}
\textbf{Minigrid}
consists of two distinct custom environments: instances of the Minigrid-SimpleCrossing and Minigrid-LavaCrossing environments. In both scenarios, the agent (represented by the red arrow) must reach the green goal square within a grid world, starting each episode at the grid square (1,1). The first environment is a $13 \times 13$ grid featuring walls along its borders and at various points in the middle. 
The presence of goals behind some walls may lead to diverse optimal plans from the agent's starting position. The second environment is a 9x9 grid with two lava crossings in the middle. If the agent touches the lava, it incurs a negative reward and a termination of the episode. In both environments, the average number of steps taken per episode is around ten steps.

\textbf{PointMaze}
We develop two distinct custom environments: PointMaze-Obstacle and PointMaze-FourRooms. In both scenarios, the agent, represented by a 2-DoF ball force-actuated in the Cartesian directions (x and y), must reach a goal within a closed maze, starting each episode from the top-left corner. The PointMaze-FourRooms environment is a classic setup featuring four rooms, with walls along its borders and various locations throughout the grid. 
Goals in the bottom-right room may lead to diverse optimal plans originating from the agent's starting position. In the PointMaze-Obstacle environment, a square object is placed in the center of the grid, obstructing the agent's path and ensuring that each goal along the diagonal can be reached via two distinct plans. On average, the agent takes approximately 80 steps to complete an episode in both environments.

\textbf{Parking} From the environments in the highway-env package, we focus on the parking continuous control task where the ego-vehicle must park in a designated space with the correct heading. Each episode begins with the agent positioned randomly in the parking lot, adding complexity as optimal paths to the same goal can vary significantly from one episode to another. On average, an episode in this environment consists of 15 steps.

\textbf{Panda-Gym} From the collection of environments provided in the panda-gym package, we focus on the Reach continuous control environment, where the Panda arm robot must reach a specified location. The agent begins each episode at the same starting position. On average, each episode in this environment consists of approximately 20 steps.

\end{document}